\documentclass[10pt,conference,a4paper]{IEEEtran}
\usepackage{times}
\usepackage{soul}
\usepackage{epsfig}
\usepackage{url}
\usepackage[hidelinks]{hyperref}
\usepackage[small]{caption}
\usepackage{graphicx}
\usepackage{amsmath}
\usepackage{booktabs}
\usepackage{makecell}
\usepackage{amssymb}
\usepackage{color, colortbl}
\definecolor{LightCyan}{rgb}{0.88,1,1}
\usepackage[shortlabels, inline]{enumitem}
\usepackage[ruled,vlined]{algorithm2e}
\newcommand{\eat}[1]{}

\ifCLASSINFOpdf
\else
\fi
\def\eg{\emph{e.g}.} 
\def\ie{\emph{i.e}.} 
 
\def\etc{\emph{etc}.}

\begin{document}
%
\title{Weakly Supervised Body Part Segmentation \\with Pose based Part Priors}

\author{\IEEEauthorblockN{Zhengyuan~Yang, Yuncheng~Li, Linjie~Yang, Ning~Zhang and Jiebo~Luo}
\author{\IEEEauthorblockN{Zhengyuan~Yang$^1$, Yuncheng~Li$^2$, Linjie~Yang$^3$, Ning~Zhang$^4$, and Jiebo~Luo$^1$}
\IEEEauthorblockA{$^1$Department of Computer Science, University of Rochester}
\IEEEauthorblockA{$^2$Google Inc., $^3$ByteDance AI Lab, $^4$Vaitl Inc.}
}
}

\maketitle

\begin{abstract}
Human body part segmentation refers to the task of predicting the semantic segmentation mask for each body part. Fully supervised body part segmentation methods achieve good performances but require an enormous amount of effort to annotate part masks for training. In contrast to high annotation costs needed for a limited number of part mask annotations, a large number of weak labels such as poses and full body masks already exist and contain relevant information. Motivated by the possibility of using existing weak labels, we propose the first weakly supervised body part segmentation framework.
The core idea is first converting the sparse weak labels such as keypoints to the initial estimate of body part masks, and then iteratively refine the part mask predictions. We name the initial part masks estimated from poses the ``part priors''.
with sufficient extra weak labels, our weakly supervised framework achieves a comparable performance ($62.0\%$ mIoU) to the fully supervised method ($63.6\%$ mIoU) on the Pascal-Person-Part dataset. Furthermore, in the extended semi-supervised setting, the proposed framework outperforms the state-of-art methods. Moreover, we extend our proposed framework to other keypoint-supervised part segmentation tasks such as face parsing.
\end{abstract}


%
\IEEEpeerreviewmaketitle

\section{Introduction}
%
%
%
%
Body part segmentation~\cite{xia2017joint,fang2018wshp} aims to decompose humans into semantic part regions with the predicted part masks. \eat{Body part segmentation is related to human parsing~\cite{gong2017look} but focuses more on predicting part masks that directly reflect body structures, \eg, ``upper/ lower arms'' instead of ``coat'' or ``jumpsuits.'' }Body part segmentation helps various vision tasks such as action recognition~\cite{zolfaghari2017chained}, person re-id~\cite{kalayeh2018human}, and image generation~\cite{Balakrishnan_2018_CVPR}. Its potential applications include but are not limited to online shopping~\cite{yamaguchi2013paper} and surveillance systems~\cite{liu2018cross}.
Previous studies on part segmentation~\cite{chen2014detect,xia2016zoom,xia2017joint,fang2018wshp} mostly follow a fully supervised setting and thus require manually annotated part masks as supervision during training. However, the number of annotations for body part segmentation is often limited due to the high annotation costs of labeling the pixel-level part masks. The Pascal-Person-Part dataset \cite{chen2014detect,xia2017joint} is the largest body part segmentation dataset by far but contains only 3.5K images in total.

On the other hand, vast amounts of available weak annotations such as human poses and full body masks already exist and contain information related to the part segmentation task. In particular, human poses~\cite{yang2011articulated,nie2018human,yang2018action,yang2018actionicpr} provide cues for high-level human structures, \ie, the coarse spatial locations of the human part. The widely used pose estimation dataset~\cite{lin2014microsoft} contains more than 30K images with 150K pose annotations. The full human mask is also widely available, as ``person'' is a semantic class in many semantic segmentation datasets~\cite{lin2014microsoft}. In this study, we focus on exploiting such abundant existing weak supervisions for part segmentation.
Our motivation of utilizing existing weak supervisions is similar to weakly supervised semantic segmentation \cite{bearman2016s,lin2016scribblesup}. However, directly applying weakly supervised semantic segmentation methods onto the part segmentation task generates poor results due to the more complicated scenes and labeling spaces. There are 17.9 instances per image on average in PASCAL-person-part compared to 2.8 object instances in PASCAL semantic segmentation.

In this study, we investigate using human poses and full masks as the alternative supervision for training the body part segmentation. The core idea is to convert the sparse human poses into an initial part mask estimates, namely the part priors, and iteratively refine the mask predictions. Specifically, the first step is to convert poses into initial part mask estimates by exploiting the prior knowledge of human structures, \eg, pixels around the connection between joints ``shoulder'' and ``elbow'' are likely to be ``upper arm.'' The uncertain boundary regions are left blank initially and are gradually recovered by an iterative refinement module. The intermediate mask predictions during the iteration are refined by the Conditional Random Field (CRF)~\cite{krahenbuhl2011efficient} and are used as the supervision to step by step improve the segmentation network. In this paper, we study how to generate and refine the part priors for weakly supervised part segmentation.

To the best of our knowledge, this is the first feasible solution on weakly supervised body part segmentation. On Pascal-Person-Part, our proposed framework achieves comparable performance ($62.0\%$ mIoU) to the fully supervised state-of-the-art method ($63.6\%$ mIoU) \cite{deeplabv3plus2018}, with additional weak supervision used. When compared in a semi-supervised setting that part masks from Pascal-Person-Part are used in training, the framework outperforms the state-of-the-art semi-supervised methods \cite{fang2018wshp}. We show that the widely available weak supervisions improve the performance of body part segmentation by a large margin. Furthermore, our part segmentation framework can be extended to other keypoint supervised segmentation tasks. We extend our weakly supervised framework onto the face parsing task on the Helen~\cite{smith2013exemplar} and AFLW dataset~\cite{koestinger11a}.

How does the proposed framework obtain significant improvements, and why can weak labels help the part parsing task? We show empirically in Section~\ref{sec:dicussion} that the successful exploitation of extra weak data, which has not been previously leveraged, is the main reason for the good performance. Qualitative results show that the gain mainly comes from reducing the error in body structure prediction, \eg, confusing arms with legs. Furthermore, extensively experiments show the superiority and robustness of the proposed pose based part prior. The new problem formulation, the significant performance improvements, and the effective mechanisms to achieve the goal, constitute the main contributions of this work.

\section{Related Works}
{\bf Human body part segmentation.} Human body part segmentation~\cite{xia2016zoom,xia2017joint,fang2018wshp} is the task of generating body part masks based on human structures.
Xia~\emph{et al.}~\cite{xia2017joint} propose to jointly conduct body part segmentation and pose estimation, and show that the two complementary tasks could help each other. Fang~\emph{et al.} \cite{fang2018wshp} propose a data augmentation method for part segmentation based on pose similarities. Part segmentation is related to human parsing \cite{Gong_2018_ECCV,gong2017look,luo2018macro,gong2018instance,nie2018mula,gong2019graphonomy,wang2019learning}, as both tasks aim to predict pixel-level semantic masks for human parts. The major difference is that body part segmentation focuses on predicting part masks that directly reflect body structures, \eg, ``upper/ lower arms.'' On the contrary, human parsing contains clothing and object classes such as ``sunglasses,'' ``hats,'' ``coats,'', \etc~ 
{\bf Weakly supervised semantic segmentation.}
Our study is also related to weakly supervised semantic segmentation. Frequently used weak supervisions for semantic segmentation include scribbles \cite{lin2016scribblesup,tang2018regularized}, bounding boxes \cite{hu2017learning, papandreou2015weakly, Remez_2018_ECCV}, points \cite{bearman2016s} and image labels \cite{kwak2017weakly, wei2017stc, zhou2016learning,kolesnikov2016seed}.
Despite the promising results achieved on the semantic segmentation task, existing weakly supervised segmentation methods require the saliency of target regions in color space. Directly applying the previous methods to more complicated scenarios, such as scene parsing \cite{Li_2018_ECCV} and part segmentation, generates limited performance. To solve the weakly supervised body part segmentation task, we use the human structure knowledge with the proposed pose based part prior.
Our study is also related to previous explorations on transforming human poses into human parts~\cite{yang2011articulated,seguin2014pose}.
\section{Methodology}
\label{sec:method}
\subsection{Overview}
\begin{figure}[t]
\begin{center}
   \centerline{\includegraphics[width=5.5cm]{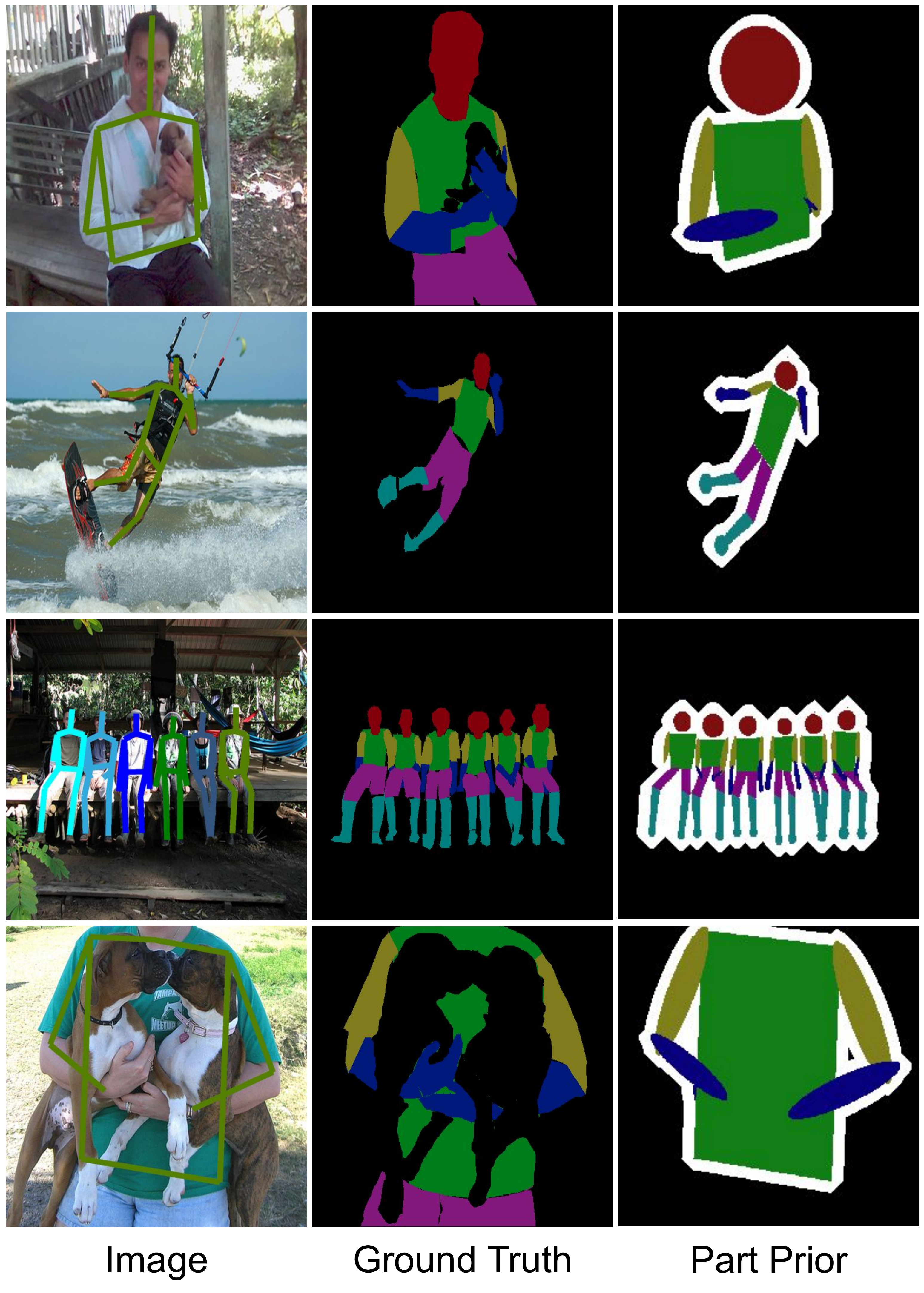}}
\end{center}
\vspace{-0.2in}
	\caption{Examples of pose based part priors on Pascal-Person-Part. The colored, black and white regions denote foreground part priors, background areas, and uncertain areas, respectively.}
\vspace{-0.1in}
\label{fig:ELL}
\end{figure}

\begin{figure*}[t]
\begin{center}
   \centerline{\includegraphics[width=14cm]{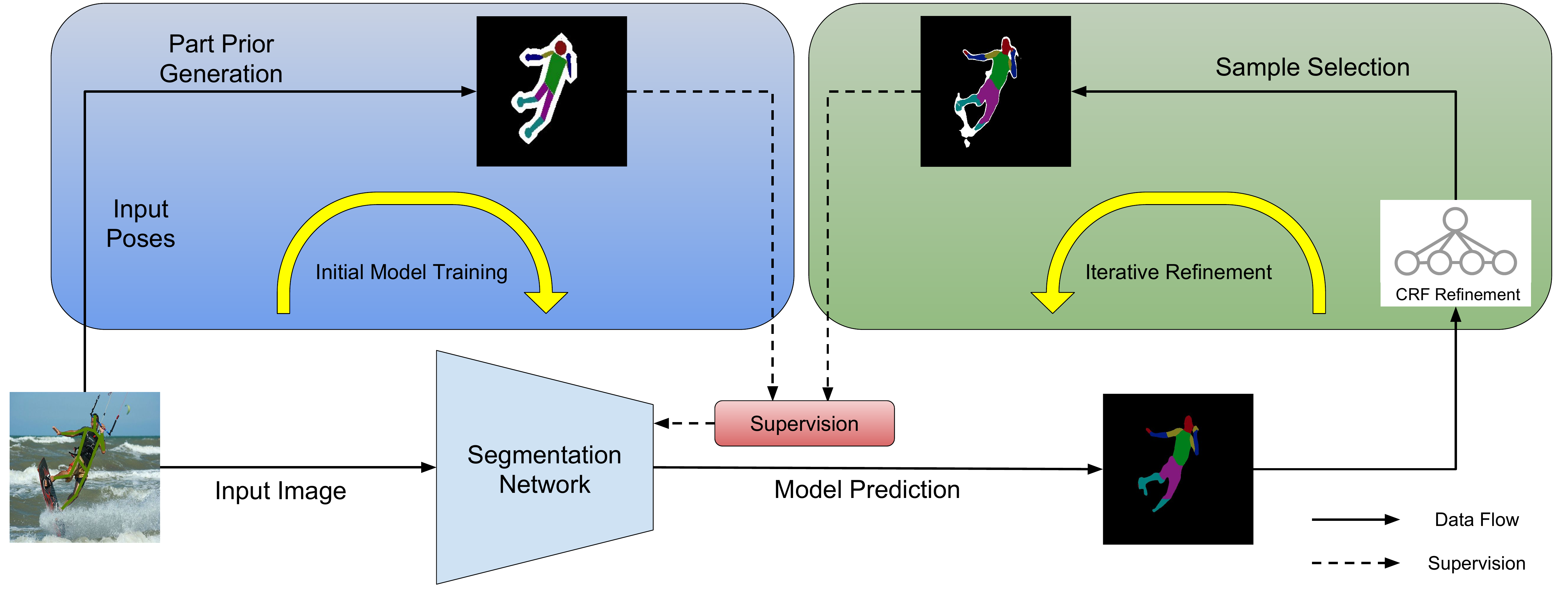}}
\end{center}
\vspace{0.0in}
	\caption{The overall framework of the proposed method. The initial model training step {\it (left blue box)} aims to learn part structures, and is supervised by the part priors generated with input poses. The iterative refinement module {\it (right green box)} then refines the detailed shapes of predicted masks with image information. In segmentation masks, the colored, black and white regions denote the foreground body parts, backgrounds and uncertain regions respectively.}
\vspace{-0.0in}
\label{fig:frame}
\end{figure*}
We first introduce the problem formulation of body part segmentation. Given an input RGB image $I_n \in \mathbb{R}^{H \times W \times 3}$ of size $H \times W$ in training set $D$ with $N$ samples, the objective of body part segmentation is to classify each pixel into one of the $C$ body part classes or background as an output mask. In fully supervised approaches, pixel-level annotations $L_n \in \left\{ 0,1,\ldots,C \right\}^{H \times W}$ are manually labeled, and the network is trained with the per-pixel cross-entropy loss $\ell$.
Such methods show good performance on tasks with sufficient pixel-level annotations. However, the number of body part annotations is still limited due to the high labeling costs. 
In this study, we investigate the approach of training body part segmentation models with weak supervisions. Capitalizing on large scale datasets such as MS-COCO~\cite{lin2014microsoft} that contain poses and full mask annotations, we train the body part segmentation framework {\it without} any pixel-level part annotations.

The overall structure of the iterative framework is shown in Figure \ref{fig:frame}. 
In order to learn dense masks with sparse keypoint coordinates, we convert the poses into part prior masks by drawing geometric shapes between the corresponding keypoint locations. The conversion rule is designed only to cover the most confident regions, \ie, to generate part priors that have high precision. Instead of learning the detailed part shapes, the objective of the part prior is to provide cues of the coarse body part location. On the estimated boundary regions, the part prior label is left blank and is later gradually recovered with iterative refinement. We use the generated prior masks to train a segmentation network with the proposed structure loss and mask loss. 
To recover the uncertain regions and get more accurate mask predictions, we introduce the iterative refinement module. The initially predicted masks on the training set are refined by a CRF and are adopted as extra training labels for the next iteration. We show that the refined masks contain more shape details as CRF exploits the low-level image features. Since the CRF refinement might lead to degeneration in structure information, both part priors and the refined mask predictions are used as supervision in the iterative training. Self-paced learning is further adopted to select reliable enough predictions as supervision during iterative refinement. Furthermore, we find that end-to-end CRF refinement such as CRFasRNN~\cite{zheng2015conditional} does not work well on this task due to the missing of full supervisions.

\subsection{Pose based Part Priors}
\label{sec:partprior}
The human pose $P=\left\{ (x_j,y_j) \right\}_{j=1}^J$ is a set of keypoints on the body structure. As a sparse representation of human body structures, poses are not directly compatible with traditional fully supervised segmentation methods, where per-pixel annotations are required. Previous studies~\cite{xia2017joint,zolfaghari2017chained,fang2018wshp} have explored the conversion from sparse poses to the part prior masks. However, with poses only, it is challenging to directly generate part priors that are consistent with part boundaries reflected in images, since poses only contain body structure information with no cues about the detailed part shapes.
Due to this inherent limitation, training the network directly with such part priors \cite{zolfaghari2017chained} generates unsatisfying masks with poor shape details. In semi-supervised studies, introducing a part prior refinement step \cite{fang2018wshp} is effective, but extra manually labeled part masks are required. 

In contrast to previous studies, we design part priors to have high precision on labeled regions, and we rely on the iterative refinement module to gradually recover the remaining uncertain boundary regions.
As shown in Figure \ref{fig:ELL}, we design part masks as ellipse or polygon templates. We use different templates for the upper body and lower body to better exploit the human structure prior. We show in section \ref{sec:dicussion} that such part priors well preserve the structure information, and do not require extra labels other than the poses.

{\bf Ellipse part priors.} We draw ellipse $E(x,y,a,b,\alpha)$ on the estimated {\it head}, {\it upper arm} and {\it lower arm} regions based on the two corresponding poses $P_i$ and $P_j$ with the ellipse parameters:
\begin{equation}
\label{equ:ell}
\begin{aligned}
& x = (P_j^x+P_i^x)/2 \quad y = (P_j^y+P_i^y)/2\\
& a = c_a * d(P_i, P_j) \quad b = c_b * d(P_i, P_j)\\
& \alpha = \arctan\left(\left(P_j^y-P_i^y\right)/\left(P_j^x-P_i^x\right)\right),
\end{aligned}
\end{equation}
where $d$ is the $L2$ distance between the two keypoints and $(x,y)$ is the center of the generated ellipse. $c_a$ and $c_b$ are pre-defined scale factors. Typical parameters for $c_a$ and $c_b$ are $0.6$ and $0.22$. Although the sizes and shapes of body parts vary, experiments in section \ref{sec:dicussion} show that the final results are robust to the hyper-parameter selection. The keypoints on the {\it shoulder} and {\it elbow}, the {\it elbow} and {\it wrist}, the {\it neck} and {\it head}, are paired to generate the ellipse part priors for the {\it upper arm}, {\it lower arm}, and {\it head}, respectively. 

{\bf Polygon part priors.}
We draw polygons on the {\it torso}, {\it upper leg} and {\it lower leg} regions as part priors. The {\it torso} prior is generated by connecting the {\it left/right shoulder} and {\it left/right hips} to generate a quadrilateral. For the {\it leg} regions, the width of the leg is estimated to be half (\textit{upper leg}) or one third (\textit{lower leg}) of the distance from the {\it left/right hip} to the hip center. We then shift the keypoints of the {\it hip} and {\it knee} or the {\it knee} and {\it ankle} horizontally left and right to generate four points based on the estimated width, and generate the quadrilateral part priors for the {\it upper leg} or {\it lower leg}. 

{\bf Estimated foregrounds.}
Furthermore, we generate estimated foreground masks based on part priors with the image dilation technique. The estimated foreground contains the generated part priors and uncertain blank regions. The size of the dilation matrix is proportional to the height of the poses. As shown in Figure \ref{fig:ELL}, the generated part priors $y$ contain three regions: estimated background $B$ indicated by black, uncertain blank regions $U$ indicated by white, and estimated foreground regions $F$ indicated by other colors.

{\bf Missing keypoints.}
Although ground-truth poses are used for part prior generation, certain keypoints can still be missing because of viewpoints or being out of image boundary. We design recovery rules based on poses' spatial priors to alleviate this problem. For example, we estimate the location of the {\it neck} as the middle point of left and right {\it shoulder}. We draw the quadrilateral part priors for the {\it torso} as a parallelogram when one of four required keypoints is missing. Although the recovery step can not completely solve the inherent limitation of incomplete poses, it provides more part prior information and improves the final performance.

{\bf Overlapping and occlusion.}
The other inherent limitation of generating part priors with sparse keypoints is overlapping and occlusion. 
First, each individual's body parts overlap due to the viewpoint and pose structures. With only 2D keypoint information, it is impossible to perfectly recover the depth order. We make such an assumption that is correct in most cases empirically: the lower arms/legs are in front of the upper arms/legs, the arms are in front of the legs, and the limbs are in front of the torso and head. 
Second, there also exist occlusions among different people. We rank the order with the number of visible keypoints, \ie, we assume the person with the most annotated keypoints is in the front of others.

In summary, the scale, location, and orientation of generated parts are calculated based on corresponding keypoints. We do not try to generate part priors that fit perfectly with the image, as this cannot be accomplished. Instead, the objective is to learn a reliable initial prediction, which provides cues for part structures. The simple design also allows for an easy adaptation to other part segmentation tasks, as shown in section \ref{sec:face}. We introduce the initial network training together with iterative refinement in section \ref{sec:iter}.

\subsection{Training Objective and Iterative Refinement}
\label{sec:iter}
In this section, we first discuss the training objective in both modules. We then introduce the iterative refinement module.

{\bf Training objective.} In each training iteration, the objective is to minimize the loss function $L$ by learning parameters $\theta$ in the segmentation model $f(I;\theta)$. The weakly supervised loss function $L$ consists of the structure loss $L_{s}$ and the full mask loss $L_{m}$. The objective function is:
\begin{equation}
\label{equ:opti}
\displaystyle \min_{\theta} \sum_{I \in D} L_{s}(f(I;\theta))+L_{m}(f(I;\theta))*w_m.
\end{equation}
In part structure learning stage with the pose generated part priors, we propose the structure loss $L_{s}$ in the format of partial cross-entropy loss.
$L_{s}$ is only calculated on the confident foreground and background regions, i.e. $F \cup B$:
\begin{equation}
\label{equ:ls}
L_{s} = \sum_{i \in F \cup B} \sum_{c=0}^C \ell (f_{i}(c), y_{i}(c)) ,
\end{equation}
where $f_i(c)$ is the model prediction score for class $c$ at pixel $i$, $y$ is the part prior generated from poses and $\ell$ is the per-pixel cross-entropy loss. With incomplete but highly confident priors on each body part, the network generates initial predictions with good part structures. 

The full human mask also contains information related to part segmentation and is widely available. Furthermore, several datasets~\cite{lin2014microsoft,gong2017look} contain large-scale existing annotations for both poses and full masks. Because of this, a full mask loss $L_{m}$ is proposed to utilize the extra weak supervision. With full mask annotation $M$, a binary cross-entropy mask loss is calculated on the whole image:
\begin{equation}
\label{equ:lm}
L_{m} = \sum_{i \in I} \sum_{fg=0}^1 \ell(\hat{M}_{i}(fg), M_{i}(fg)),
\end{equation}
where foreground-background prediction $\hat{M}_i$ is generated by the predicted background class probability $f_{i}(0)$.

{\bf Iterative refinement.} 
To recover the blank uncertain regions and get better mask details, we propose the iterative refinement module.
Although poses do not contain shape information, we show that the mask shapes can be inferred from low-level pixel similarities. In the proposed iterative refinement module, we first refine the initial part prior supervised mask predictions on the training set with CRF. We adopt the dense CRF~\cite{krahenbuhl2011efficient} in the iterative refinement step, where an appearance kernel and a smoothness kernel are used as the pairwise term.
The refined masks show better prediction details by inferring pixel RGB similarities in the CRF. However, the downside of the refined masks is that the part structure information can be mistaken. For example, the predicted mask of \textit{arm} might be incorrectly labeled as \textit{torso} after the CRF refinement. 
Therefore, we propose to train the network jointly with initial part priors and generate new masks to include both structure and detail information in the iterative refinement.

Because of the variances in the complexity of the scene in image $I$ and the quality of the corresponding predicted masks $\hat{y}_{i}$, different image label pairs contribute unequally towards segmentation model learning. Incorrectly predicted masks become noises in the next iteration and therefore deteriorate the model performance. Inspired by self-paced learning, we alleviate the problem with training sample selection during the iterative refinement. We follow a previous study \cite{you2015robust} to discard unreliable predictions with a probability $p_i$ in Eq. \ref{equ:pi}, and skip samples that have low prediction confidences in the next training iteration:
\begin{equation}
\label{equ:pi}
p_i = max(0, 2-\exp(\overline{f}))
\end{equation}
where $\overline{f}$ is the averaged pixel-wise prediction confidence over foreground regions:
\begin{equation}
\overline{f} = \frac{1}{N_F} \sum_{i \in F} \left( \max_{c=1\cdots C} f_{i}(c) \right)
\end{equation}
\section{Experiments}
In this section, we first introduce the datasets used for experiments. We then compare the proposed body part segmentation methods to other state-of-the-art methods. Extensive discussions are conducted on various aspects of the method. Finally, we show the extension of the framework to other point supervised part segmentation tasks such as face parsing. The proposed framework is generally applicable for many segmentation methods, and we adopt Deeplab \cite{deeplabv3plus2018} as the base segmentation model in this study.
\subsection{Datasets}
\begin{table*}[t]
\centering 
\caption{The body part segmentation performance on Pascal-Person-Part compared to the state-of-the-art. The best weakly supervised performances are highlighted with bold and the best fully supervised performances are marked with underlines. We show the weakly supervised performance with and without using extra data from MSCOCO. More discussions regarding the data size are in Table \ref{table:size}.}
\vspace{-0.0in}
\begin{tabular}{ c c c | c c c c c c c}
    \hline
    Methods & Supervision & mIoU & Head & Torso & U-Arm & L-Arm & U-Leg & L-Leg & Bkg\\
    \hline
	LIP \cite{gong2017look} & Fully &  59.36 & 83.3 & 62.4 & 47.8 & 45.6 & 42.3 & 39.5 & 94.7 \\
	LG-LSTM \cite{liang2016semantic} & Fully & 57.97 & 82.7 & 61.0 & 45.4 & 47.8 & 42.3 & 38.0 & 88.6 \\
	Graph LSTM \cite{liang2016semantic-graph} & Fully & 60.16 & 82.7 & 62.7 & 46.9 & 47.7 & \underline{45.7} & 40.9 & 94.6 \\
	DeepLab \cite{deeplabv3plus2018} & Fully & \underline{63.64} & \underline{84.6} & \underline{66.9} & \underline{56.0} & \underline{54.2} & 45.5 & \underline{43.4} & \underline{94.9}\\
	\hline
	Ours (Part prior) & Weakly & 40.91 & 54.9 & 35.1 & 34.2 & 32.4 & 19.0 & 24.5 & 86.2\\
	FastNet \cite{zolfaghari2017chained} & Weakly & 42.11 & 61.6 & 37.8 & 32.7 & 29.2 & 20.8 & 25.0 & 87.1\\
	Ours (Part prior supervision) & Weakly & 43.91 & 50.6 & 47.0 & 31.8 & 29.5 & 29.3 & 27.9 & 87.0\\
    \hline
	Ours (Base, only PASCAL data) & Weakly & 53.54 & 76.1 & 54.8 & 39.6 & 36.5 & 37.9 & 36.1 & 93.9 \\
	Ours (Iter, only PASCAL data) & Weakly & 54.72 & 76.7 & 55.6 & 40.2 & 38.4 & 38.6 & 38.9 & 94.6 \\
	Ours (Base, with COCO data) & Weakly & 60.35 & 78.3 & 59.6 & 46.6 & 45.4 & 46.6 & 50.0 & 95.8\\
	Ours (Iter, with COCO data) & Weakly & {\bf 62.05} & {\bf 79.6} & {\bf 62.0} & {\bf 48.1} & {\bf 48.5} & {\bf 48.7} & {\bf 51.8} & {\bf 95.8} \\
    \hline
\end{tabular}
\vspace{-0.1in}
\label{table:pascal}
\end{table*}
{\bf Body part segmentation datasets.} 
{\em The Pascal-Person-Part dataset} \cite{chen2014detect,xia2017joint} contains annotations for 14 human joints and 6 body parts. The annotated body parts are Head, Torso, Upper/Lower Arms, and Upper/Lower Legs. The total 3,533 images are split into 1,716 images for training and 1,817 images for testing.  
{\em The MSCOCO dataset} \cite{lin2014microsoft} contains over 150K human instance annotations with 1.7 million labeled keypoints. In this study, the pose annotations on 31K training images are used as the extra training data.
We do not use {\em the LIP dataset} \cite{gong2017look} in this study, because the extra clothing and object classes are unrelated to body part segmentation.

{\bf Face parsing datasets.} 
{\em The Helen dataset} \cite{le2012interactive} provides 194-point dense facial landmark annotations. A 10-class pixel-level annotation is defined and generated by Smith \emph{et al.} \cite{smith2013exemplar}. The dataset contains 2,000 training images and 330 testing images. We follow the split in previous studies \cite{smith2013exemplar} and evaluate the face parsing task on the Helen testing set. 

{\em The AFLW dataset} \cite{koestinger11a} contains around 25K annotated faces. Each face has 21 labeled facial landmarks. We filter out a 6K image subset that contains images with front-view faces. The AFLW dataset is used as the weakly supervised training data for the face parsing task evaluated on Helen.
\subsection{Body Part Segmentation Results}
{\bf Weakly supervised results.}
Table \ref{table:pascal} reports the body part segmentation results on the Pascal-Person-Part dataset. The {\bf top portion} of the table contains the numbers of several fully supervised state-of-the-art parsing methods~\cite{gong2017look,liang2016semantic,liang2016semantic-graph,deeplabv3plus2018}. In order to separate the improvements brought by the proposed weakly supervised framework and that by advanced network structures, we use the same network~\cite{deeplabv3plus2018} when comparing fully and weakly supervised methods. Therefore, we refer to the number generated by Deeplab~\cite{deeplabv3plus2018} as the fully supervised results in the following discussions. 
The results of the compared weakly supervised baselines are shown in the {\bf middle} of Table \ref{table:pascal}. ``Ours-part prior'' directly evaluates the generated part priors as final predictions. ``Ours-part prior supervision'' takes the generated part priors as full supervisions and train the parsing network with the per-pixel cross-entropy loss~\cite{long2015fully}. 
Finally, the four rows at the {\bf bottom} are the variations of our approach. 

When iterative refinement and extra pose annotations from MSCOCO are used, the proposed weakly supervised approach ``Ours-Iter-with COCO'' achieves a mIoU of $62.05\%$, which is fairly comparable to the $63.64\%$ acquired by the fully supervised methods \cite{deeplabv3plus2018} with the same network structure. We find there are two major reasons for the improvement. First, the proposed weakly supervised part segmentation framework makes learning from poses feasible, and improves the weakly supervised baseline performance by $10.8\%$ in mIoU (cf. ``Part prior supervision'' and ``Iter, only PASCAL data''in Table~\ref{table:pascal}). Second, the successful exploitation of extra weak supervisions further improves the performance by $7.3\%$ (cf. ``Iter, only PASCAL data'' and ``Iter, with COCO data'').

{\bf Semi-supervised results.}
The proposed framework also performs well under the semi-supervised setting. For semi-supervised learning, we replace part priors on Pascal-Person-Part with ground-truth part masks. All other settings remain the same as the weakly supervised experiment. Without learning specific adaptation networks to close the gap between different supervision types and domains, the proposed framework still generates a mIoU of $68.88\%$ that outperforms the state-of-the-art ($67.60\%$ mIoU) \cite{fang2018wshp}. 

\subsection{Qualitative results analyses}
\label{sec:qualitative}
\begin{figure*}[t]
\begin{center}
   \centerline{\includegraphics[width=15cm]{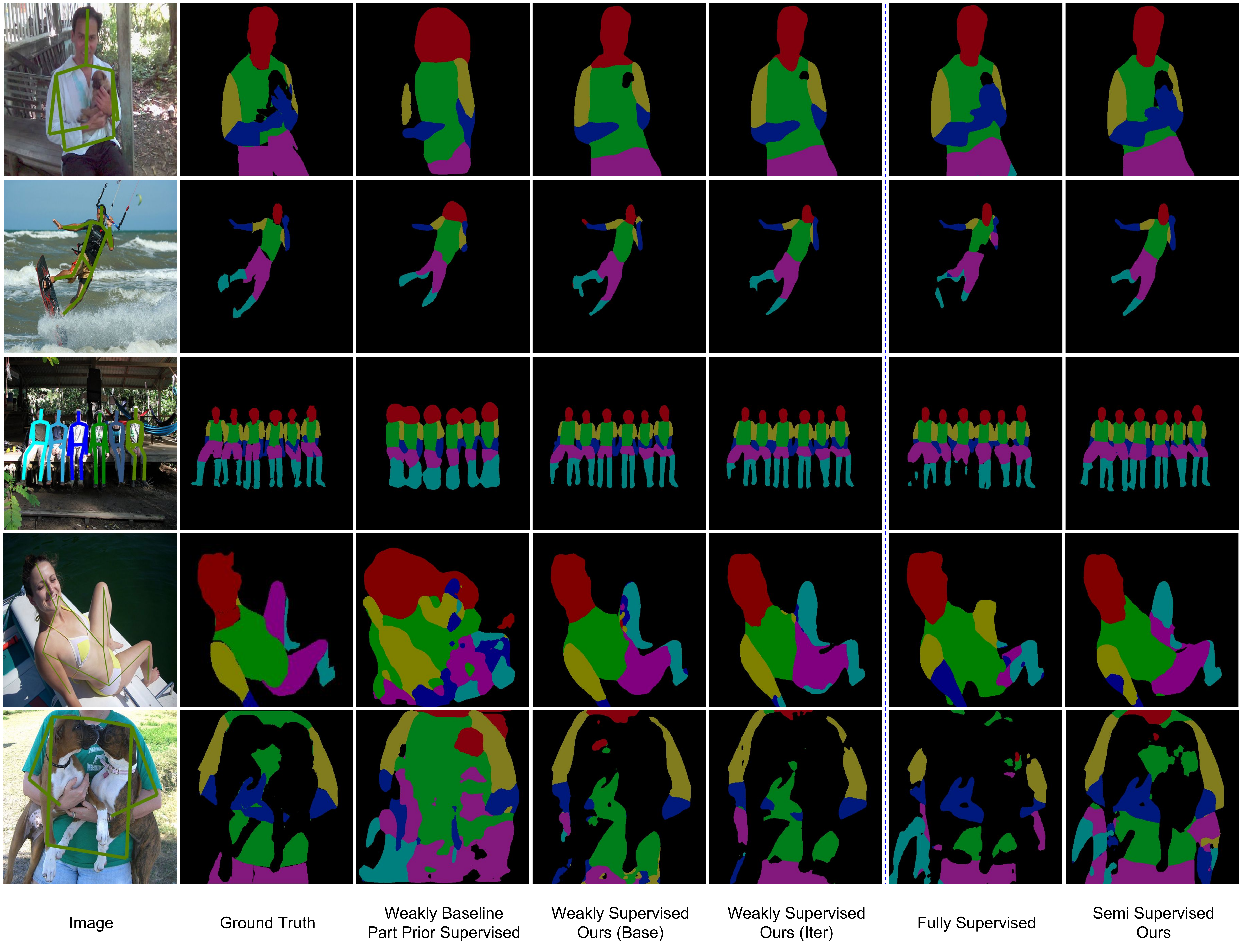}}   
\end{center}
\vspace{-0.2in}
	\caption[]{The qualitative results on Pascal-Person-Part. The middle three columns are weakly supervised results. The right two columns utilize part segmentation annotations, i.e. fully and semi-supervised. Additional qualitative results can be found in the following link\footnotemark.}
\vspace{-0.0in}
\label{fig:pascal}
\end{figure*}
In this section, we analyze the success and failure cases of our model as well as the compared methods to show the advantages and limitations of the proposed weakly supervised parsing framework. Qualitative results are shown in Figure \ref{fig:pascal}. The first column is the input image with pose annotations visualized. The second to the forth columns show the ground-truth masks, weakly supervised baseline ``Ours-part prior supervision'', our weakly supervised method ``Ours-Base-with COCO'' and ``Ours-Iter-with COCO'' respectively. The right two columns include the fully supervised results~\cite{deeplabv3plus2018} and our semi-supervised results.

We show empirically from the qualitative results that the major advantage of using weak supervisions is to generate better part structures. For example in the second row, our weakly supervised method correctly generates masks for the {\it lower leg} regions, which are mispredicted as {\it background} by the fully supervised method. Similar examples can be observed in the second, third, and forth row of Figure \ref{fig:pascal}. This observation is also applicable to semi-supervised results.

We then show the effectiveness of the iterative refinement.
As shown in the fifth column, the refined results contain more accurate local details compared to the ones without iterative training in the fourth column. One clear example is the \textit{neck} regions in the first row, which now have more precise boundaries. 

\begin{figure}[t]
\vspace{-0.1in}
\begin{center}
    \centerline{\includegraphics[width=7.5cm]{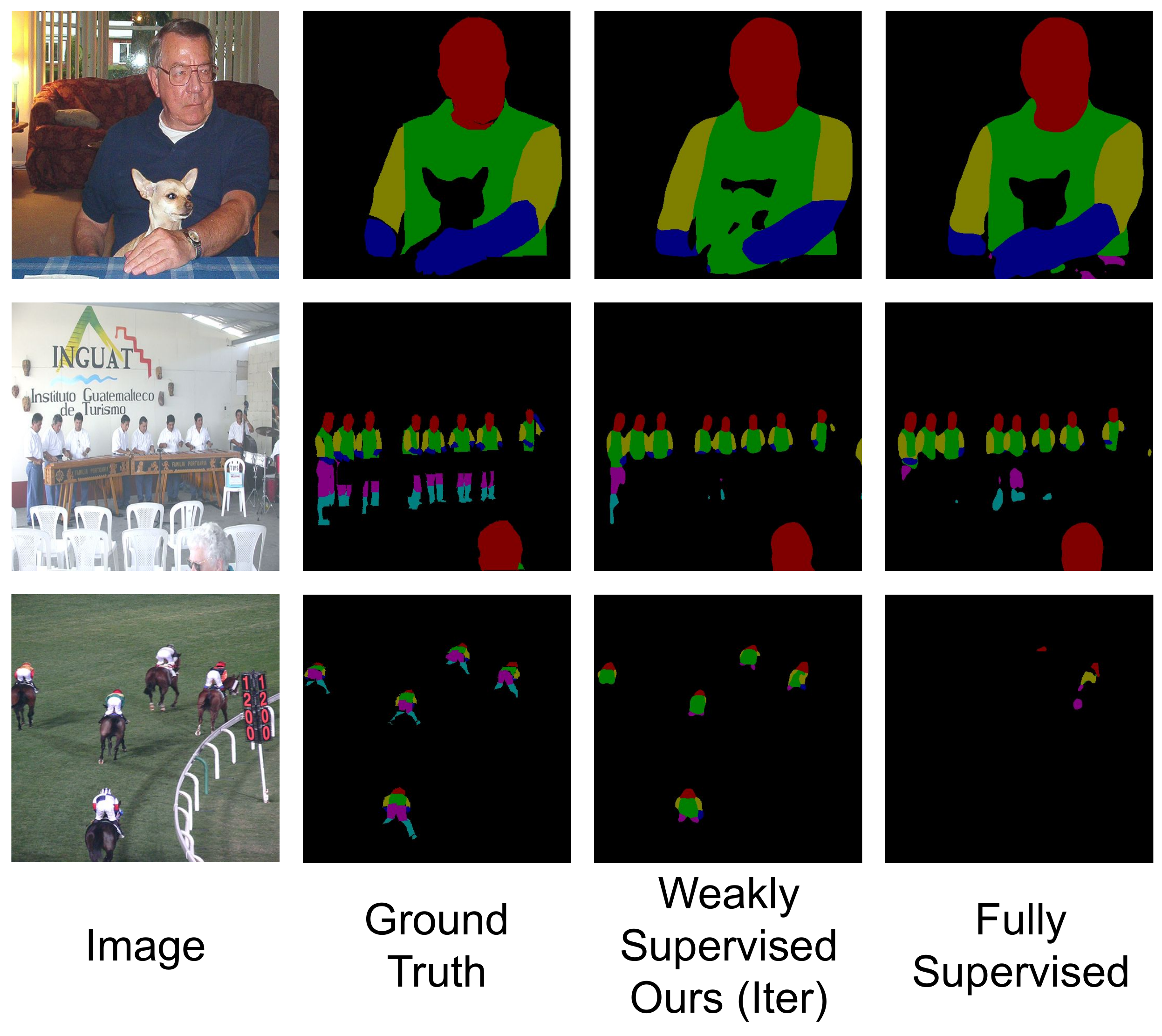}}
\end{center}
\vspace{-0.2in}
	\caption{The common failures of our weakly supervised method.}
\vspace{-0.2in}
\label{fig:fail}
\end{figure}

{\bf Failure cases of our approach.} 
Figure~\ref{fig:fail} shows the failure cases of our model. We observe three types of common failures: 1) The model might incorrectly predict occluded regions as one of the body parts instead of the background. For example, the dog in the first row of Figure~\ref{fig:fail}. 2) Part priors can not be generated when corresponding joints are occluded or out of image boundaries. Because of this, such regions might be mispredicted as background. For example, the {\it leg} regions in the second row. 3) Challenging cases with tiny body parts or complicated scenes might also fail the model, \eg, the third row of Figure~\ref{fig:fail}. \footnotetext{Additional qualitative results:\\ \url{http://cs.rochester.edu/u/zyang39/weakly_parsing/pascal_visu.html}}

\subsection{Discussions}
\label{sec:dicussion} 
\begin{table}[t]
\centering
\caption{Comparison on different part prior generation methods. ``Our part prior, w recovery'' is our results to be compared to.}
\vspace{-0.0in}
\begin{tabular}{c c c}
    \hline
    Methods & mIoU &  \\
    \hline
    \makecell{Skeleton label map \cite{xia2017joint}} & 50.34 &  \\
    \hline
    Our part prior, w/o recovery & 52.77 &  \\
    Our part prior, w recovery & 54.72 &  \\
    \hline
    Our part prior, large param. & 51.94 &  \\
    Our part prior, small param. & 52.37 &  \\
    Our part prior, ideal param. & 58.31 &  \\
    \hline
\end{tabular}
\vspace{-0.2in}
\label{table:prior}
\end{table}
{\bf Different part prior generation methods.}
We compare different part prior generation methods in Table \ref{table:prior} to show the superiority and robustness of the proposed part prior. In comparison, we report the performance of iterative training with only Pascal-person-part data. In the {\bf top} portion of Table \ref{table:prior}, we compare our part prior generation methods to the skeleton label map~\cite{xia2017joint}. The proposed part prior generation method ($54.72\%$ mIoU) outperforms the skeleton label map ($50.34\%$ mIoU)~\cite{xia2017joint}, which draws a stick between neighboring keypoints. 
In the {\bf middle} of the table, we validate the effectiveness of overlapping recovery introduced in section \ref{sec:partprior}. The proposed recovery method improves mIoU by $2\%$.

With various body part sizes and shapes, it is impossible to select hyper-parameters that generate perfect part priors. At the {\bf bottom} of the table, we show empirically that the framework is robust against different shape hyper-parameters. The upper bound of hyper-parameter selection is shown in ``Our part prior, ideal parameters,'' where ground-truth masks are used to fit the hyper-parameters. In the ``large parameters'' and ``small parameters'' experiment, we increase and decrease the shape hyper-parameters by $50\%$, respectively. By including the uncertain regions in part priors, the framework is robust against poor hyper-parameters, and the results remain better than other part prior generation methods~\cite{xia2017joint}.
\begin{table}[t]
\centering 
\caption{Weakly and semi supervised body part segmentation performance, trained with different amounts of extra weak annotations.}
\vspace{-0.0in}
\begin{tabular}{ c c c c}
    \hline
    Methods & Pascal & COCO & mIoU \\
    \hline
    Ours (Weakly) & 1.7K & 0K & 54.72 \\
    Ours (Weakly) & 1.7K & 10K & 59.72 \\
    Ours (Weakly) & 1.7K & 31K & 62.05\\
    \hline
    Fully & 1.7K (part masks) & 0K & 63.64 \\
    Ours (Semi) & 1.7K (part masks) & 10K & 67.68 \\
    Ours (Semi) & 1.7K (part masks) & 31K & 68.88 \\
    \hline
\end{tabular}
\vspace{-0.0in}
\label{table:size}
\end{table}

{\bf The influence of training data size.}
As shown in Table \ref{table:size}, we evaluate the influence of data size and supervision types. We train the proposed framework with different amounts of full annotations, weak annotations, or a mix of both. In fully- and semi-supervised experiments, we use the part mask annotations on Pascal-Person-Part that contain 1.7K images in total. In weakly supervised experiments, we use the poses and full masks from MSCOCO and Pascal-Person-Part. There are 31K images with poses and full masks on MSCOCO and 1.7K on Pascal-Person-Part. We evaluate all experiments on the testing set of Pascal-Person-Part.
In Table \ref{table:size}, we compare the weakly and semi supervised performances when using \begin{enumerate*}[1)]
\item no COCO data, \item a subset of 10K COCO data, \item and all available 31K COCO data. 
\end{enumerate*}
Our framework achieves significant improvements when extra weak supervision is adopted, \ie, more than {\bf 5} mIoU for both the weakly and semi-supervised settings. The good performance is achieved by the successful exploitation of extra weak annotations, which have not been previously leveraged.

\begin{figure}[t]
\begin{center}
   \centerline{\includegraphics[width=6cm]{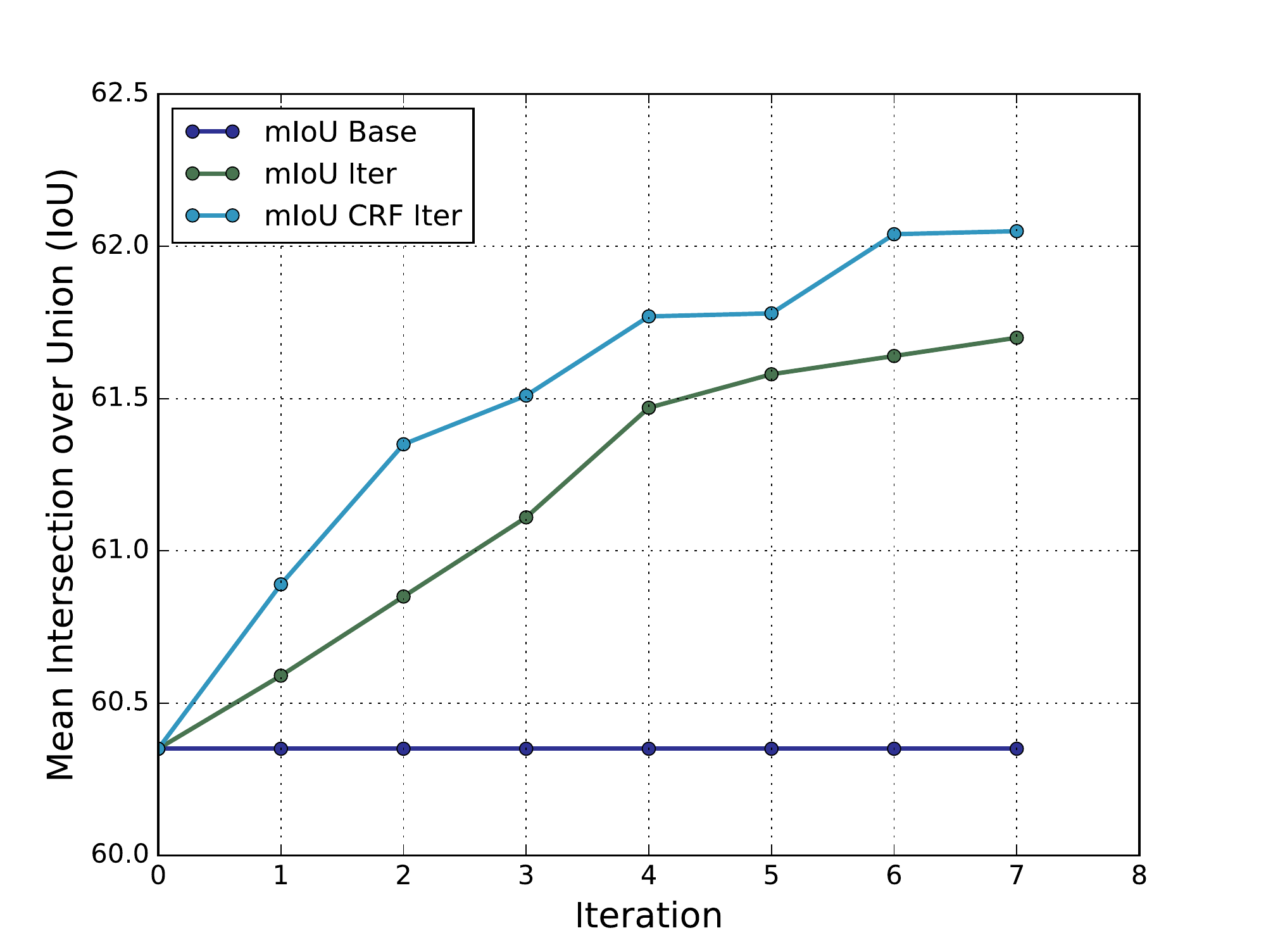}}
\end{center}
\vspace{-0.3in}
	\caption{The mIoU on Pascal-Person-Part during the iterative training.}
\vspace{-0.2in}
\label{fig:iter}
\end{figure}
{\bf Iterative training.}
We show the effectiveness of iterative training in this section. As shown in Figure \ref{fig:iter}, iterative training brings steady improvements in mIoU during the first few iterations. As expected, observations show that the improvements are mainly achieved by generating better predictions on boundary pixels. Besides, conducting CRF refinement during iterative training brings an extra gain in performance.

{\bf Training without foreground masks.}
Besides using both poses and full masks, we show that the proposed framework also generates a good performance with only pose supervisions. Without full masks, the model achieves a mIoU of $53.62\%$, which is significantly better than the weakly supervised baselines ($43.91\%$ mIoU). Under the semi-supervised setting, the proposed framework achieves a mIoU of $66.68\%$.

\subsection{Extensions to other Part Segmentation Tasks}
\label{sec:face}
Furthermore, the proposed pipeline can be easily adapted to other point-supervised part segmentation tasks, such as hand parsing, face parsing, and general object part segmentation. Such extensions could benefit other human-centered tasks such as emotion recognition~\cite{yang2019human}. As an example, we evaluate the framework on face parsing. 
In the weakly supervised face parsing task, we adopt the sparse 21 point facial landmarks as supervision. 
Similar to body part segmentation, we convert landmarks into part priors by drawing polygons. 
The facial landmark annotations from AFLW \cite{koestinger11a} are used to train the face parsing model. The face part definition follows previous studies \cite{smith2013exemplar} with two keypoints unrelated classes {\it lips} and {\it hair} merged. The method is evaluated on Helen~\cite{le2012interactive,smith2013exemplar} that has dense face parts annotations.

\begin{figure}[t]
\begin{center}
   \centerline{\includegraphics[width=8cm]{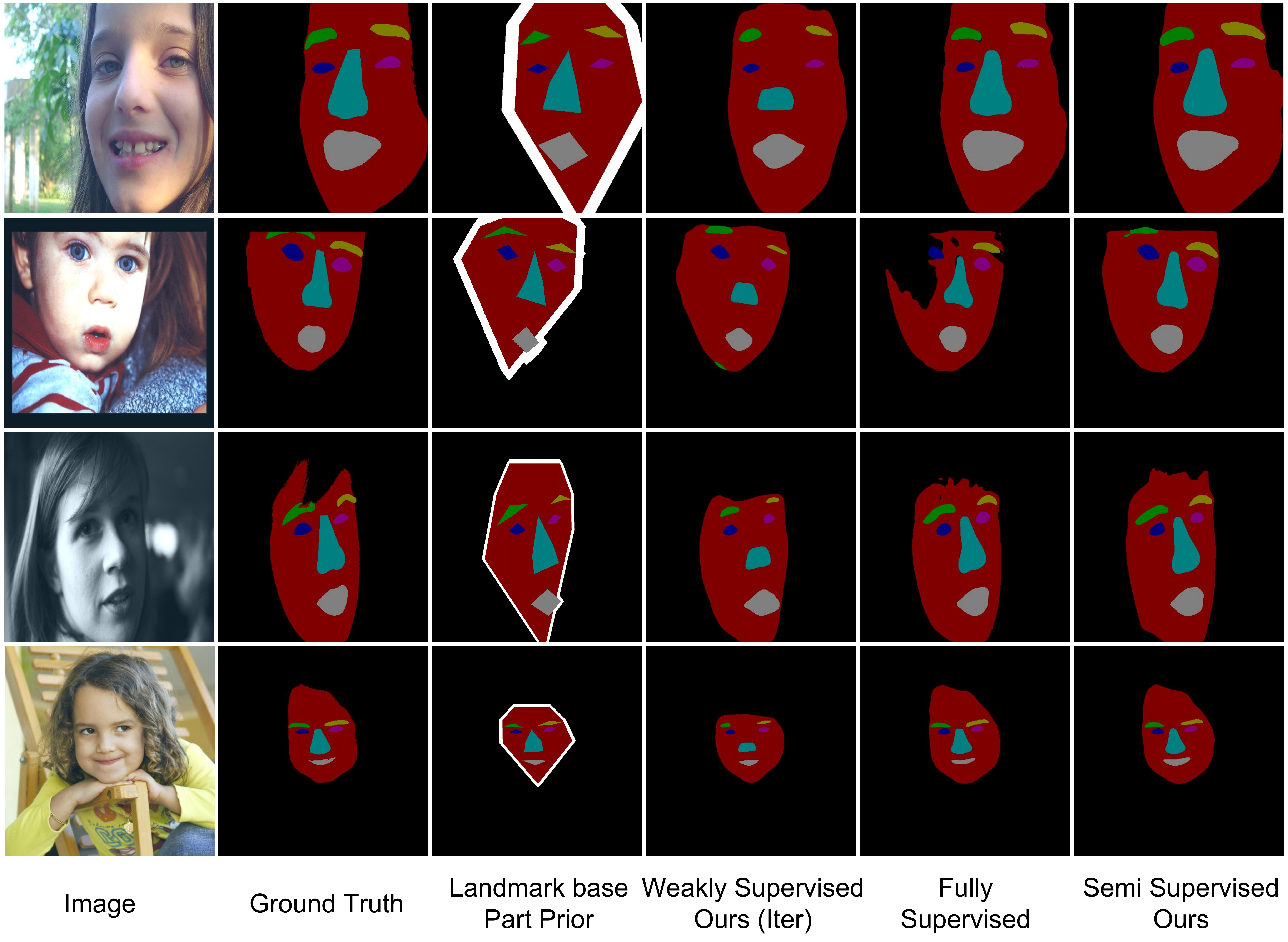}}
\end{center}
\vspace{-0.2in}
	\caption{The qualitative results on the Helen dataset.}
\vspace{-0.1in}
\label{fig:helen}
\end{figure}

As shown in Table \ref{table:helen}, the proposed framework achieves a large improvement compared to the weakly supervised baseline. Similar to body part segmentation, the semi-supervised results with more data outperform the fully supervised baseline by a large margin. Figure \ref{fig:helen} shows the qualitative results. The results show that the proposed framework can effectively utilize the abundant existing weak supervision to learn models when no full supervision is available, or combine with a small number of full annotations to further improve the performance.
\begin{table}[t]
\centering 
\caption{The face parsing performance on the Helen dataset.}
\vspace{-0.0in}
\begin{tabular}{ c c c}
    \hline
    Methods & Supervision & mIoU \\
    \hline
    Baseline (Part Prior Supv.) & Weakly & 42.54 \\
    Ours (Base) & Weakly &  47.21 \\
    Ours (Iter) & Weakly & 54.05 \\
    \hline
    Fully Supervised & Fully & 68.48 \\
    Ours (Semi) & Semi & 73.95 \\
    \hline
\end{tabular}
\vspace{-0.1in}
\label{table:helen}
\end{table}

\section{Conclusion}
To harvest the existing abundant weak annotations, we propose the first weakly supervised body part segmentation framework.
The framework first uses pose-based part priors to learn the coarse locations of each part. The iterative refinement module then gradually recovers predictions on uncertain regions. The proposed framework works well in both weakly and semi-supervised settings by effectively using the extra weak annotations, which have not been previously leveraged. The weakly supervised performance with sufficient weak annotations is comparable to the fully supervised results with the same network, while the semi-supervised results outperform the state-of-the-art. Furthermore, we show that the framework can be applied to other weakly supervised part segmentation tasks with promising results on face parsing.

\section*{Acknowledgment}
This work is supported in part by NSF awards IIS-1704337, IIS-1722847, and IIS-1813709, Twitch Fellowship, as well as our corporate sponsors.





%
\bibliographystyle{IEEEtran}
\bibliography{ieee}
\end{document}